\documentclass{article}
\usepackage{spconf,amsmath,amssymb,graphicx,setspace}
\usepackage{bm}
\usepackage{cite}
\usepackage{url}
\usepackage{float}
\usepackage{booktabs}
\usepackage{makecell}
\usepackage{multirow}
\usepackage{placeins}
\usepackage[table]{xcolor}
\definecolor{oursbg}{HTML}{FCF0E4}

\newcommand{\UnifiedTableStyle}{%
    \fontsize{8.2}{8.8}\selectfont
    \renewcommand{\arraystretch}{1.0}%
    \setlength{\tabcolsep}{3.5pt}%
    \setlength{\aboverulesep}{0.2ex}%
    \setlength{\belowrulesep}{0.2ex}%
    \setlength{\cmidrulesep}{0.1ex}%
}

\title{LATENT DATASET DISTILLATION FOR HUMAN MOTION PREDICTION}

\name{Ge Tian \quad Guang Li$^{*}$ \quad Takahiro Ogawa \quad Miki Haseyama \thanks{
This research was supported in part by JSPS KAKENHI Grant Numbers JP24K02942 and JP25K21218. Correspondence to Guang Li \texttt{<guang@lmd.ist.hokudai.ac.jp>}.}}
\address{Hokkaido University \\
    \{tian, guang, ogawa, mhaseyama\}@lmd.ist.hokudai.ac.jp}

\begin{document}
\ninept
\maketitle

\begin{abstract}
Dataset distillation (DD) compresses a large training set into a compact synthetic set while preserving downstream training utility. Although DD has been widely studied for images and recently extended to time-series forecasting, its application to human motion prediction remains largely unexplored. Human motion is high-dimensional and structurally coupled, and gradient matching (GM) in the original motion space optimizes many correlated variables without a prior on pose plausibility or temporal dynamics, which frequently yields implausible and unstable synthetic motions. To address this limitation, we propose a latent DD framework that regularizes distillation with a learned motion prior. Motions are first compressed by a residual-quantized variational autoencoder (RVQ-VAE), and distillation then updates only a learnable latent bank through the frozen quantizer and decoder. The pretrained decoder restricts synthetic motions to its output space, while residual quantization progressively refines the latent approximation across multiple codebooks and alleviates the representational bottleneck of single-stage vector quantization. Experiments on Human3.6M, CMU, and 3DPW with two prediction backbones show that the proposed framework outperforms direct GM in 27 of 30 evaluated settings and random subsets in every setting, and produces visibly more plausible synthetic motions in qualitative comparisons.
\end{abstract}
\begin{keywords}
Dataset Distillation, Human Motion Prediction, Gradient Matching, RVQ-VAE
\end{keywords}

\section{Introduction}

Dataset distillation (DD) replaces a large training set with a compact synthetic set that preserves downstream training utility~\cite{wang2018dataset, li2022awesome}. Gradient matching (GM) synthesizes such data by aligning the parameter gradients induced by real and synthetic samples~\cite{zhao2021gradient}, and later studies improve condensation through differentiable augmentation~\cite{zhao2021dsa}, distribution matching~\cite{zhao2023dm}, trajectory matching~\cite{cazenavette2022mtt}, synthetic-data parameterization~\cite{kim2022idc}, and representative matching~\cite{liu2023dream}. The resulting synthetic sets have enabled applications such as privacy-preserving medical data sharing~\cite{li2022gastric}. All of these methods were developed for images, where a synthetic sample is a freely optimized pixel array with a fixed layout.

DD has recently been extended to temporal data through frequency and trajectory matching~\cite{miao2024timedc}, harmonic matching for forecasting~\cite{hong2026hdt}, and spatio-temporal condensation~\cite{kwon2026stemdist}. Human motion, however, differs fundamentally from generic time series. Each frame is a pose whose parameters are coupled by the kinematic structure of the body, and motion predictors model the temporal dependencies among these parameters using recurrent~\cite{martinez2017motion}, graph-based~\cite{mao2019trajectory,li2020dmgnn}, attention-based~\cite{mao2020attention,aksan2021sttransformer}, and MLP~\cite{guo2023simlpe} architectures. Training such predictors is routinely repeated across architectures and horizons, so a compact synthetic set that retains training utility is of practical interest.

Despite this potential, applying GM directly to motion data is problematic. Direct GM treats every value of a synthetic sequence as an independent variable and updates it using gradient signals from the predictor alone. For articulated motion, this means searching over many strongly correlated rotation parameters without a prior on which joint configurations and temporal transitions are plausible. Distilled motions then exhibit implausible joint configurations and abrupt transitions, training becomes unstable, and the resulting sets can be inferior even to random subsets.

A natural solution is to regularize distillation with a learned motion prior rather than optimizing freely in the original space. Latent-space DD reduces the optimization burden of synthetic data for images~\cite{duan2023latent}, and discrete representations learned by VQ-VAE~\cite{oord2017vqvae} are effective for motion modeling~\cite{zhang2023t2mgpt}. A single codebook, however, imposes a restrictive bottleneck that limits how well a synthetic set adapts to the distillation objective, whereas residual vector quantization (RVQ) represents a latent vector as a sum of codewords drawn progressively from multiple codebooks and relieves this bottleneck while retaining a structured motion prior~\cite{guo2024momask}.
Building on this observation, we pretrain an RVQ-VAE motion autoencoder and distill in its pre-quantization latent space, updating only a learnable latent bank by GM through the frozen quantizer and decoder, and evaluate the framework on three benchmarks with two prediction backbones.

\begin{figure*}[t]
    \centering
    \includegraphics[width=1\textwidth]{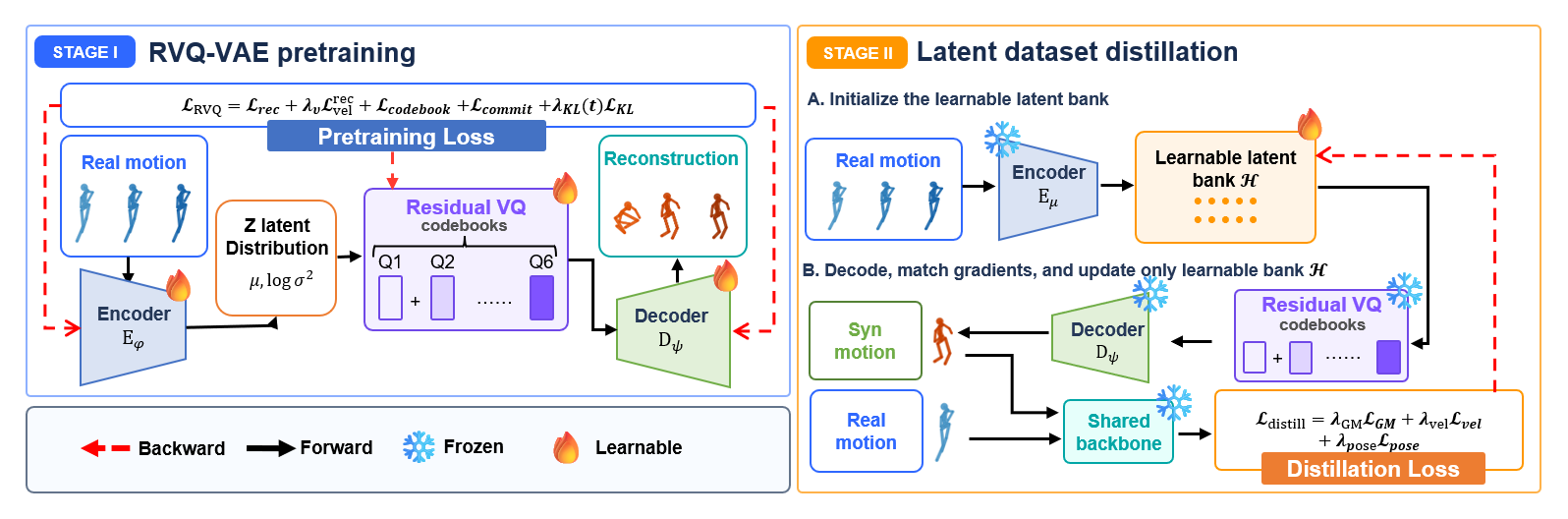}
    \caption{An illustration of the proposed latent dataset distillation framework. Stage I learns a compact motion representation with an RVQ-VAE. In Stage II, real segments are encoded to initialize a learnable latent bank $\mathcal{H}$, which is the only quantity optimized while the RVQ-VAE remains frozen.}
    \label{fig:method}
\end{figure*}

Our contributions are summarized as follows.

\begin{itemize}
\item We study dataset distillation for human motion prediction and show that direct GM in the original motion space, lacking a learned motion prior, yields implausible synthetic motions.
\item We propose a latent distillation framework built on a pretrained RVQ-VAE, in which only a learnable latent bank is optimized by GM through the frozen residual quantizer.
\item We show empirically that, under a shared objective, distilling in the residual-quantized latent space outperforms direct GM as well as continuous and single-stage latent variants.
\end{itemize}

\section{Methodology}
\label{sec:method}

\subsection{Preliminaries}

\textbf{Gradient matching.} GM constructs synthetic data by matching the parameter gradients of real and synthetic samples. A sample is divided into an observed sequence $P\in\mathbb{R}^{T_{\mathrm{in}}\times D}$ and a future sequence $F\in\mathbb{R}^{T_{\mathrm{out}}\times D}$, where $D$ is the dataset-specific feature dimension. Given a backbone $f_{\theta}$ and prediction objective $\ell$, the gradients induced by real and synthetic samples are $g_r=\nabla_{\theta}\ell(f_{\theta}(P_r),F_r)$ and $g_s=\nabla_{\theta}\ell(f_{\theta}(P_s),F_s)$. We minimize the normalized discrepancy between the two,
\begin{equation}
\mathcal{L}_{\mathrm{GM}}
=
\frac{
\sum_p
\left\|
g_s^{(p)}-\operatorname{sg}(g_r^{(p)})
\right\|_2^2
}{
\sum_{p'}
\left\|
\operatorname{sg}(g_r^{(p')})
\right\|_2^2+\epsilon
},
\label{eq:gm}
\end{equation}
where $p$ and $p'$ index parameter tensors and $\operatorname{sg}(\cdot)$ denotes stop-gradient. The real gradients are fixed targets, whereas the synthetic branch retains the computation graph so that the synthetic representation can be optimized. Unlike the original GM formulation, which interleaves synthetic-data updates with network training, the backbone is held fixed as a shared gradient probe.

\textbf{Motion data format.} Each motion sequence is represented as $X=[x_1,\ldots,x_T]$ with $x_t\in\mathbb{R}^{D}$, where $x_t$ is the pose at time $t$. A 35-frame window is divided into $T_{\mathrm{in}}=25$ observed and $T_{\mathrm{out}}=10$ future frames, and each synthetic trajectory contains 50 frames so that several prediction windows can be extracted from it.

\subsection{Residual-Quantized Motion Representation}

Direct GM optimizes motion features without an explicit structural prior. To regularize the optimization with a learned motion prior, we first train an autoencoder that combines a stochastic VAE encoder with residual vector quantization.

\textbf{Stochastic latent encoder.} Given a motion segment $X\in\mathbb{R}^{T\times D}$, the encoder produces the mean and log-variance of a latent distribution, $(\mu,\log\sigma^2)=E_{\phi}(X)$, from which $z=\mu+\sigma\odot\varepsilon$ is sampled with $\varepsilon\sim\mathcal{N}(0,I)$. Temporal convolutions reduce the temporal resolution and map each latent step to a 64-D vector.

\textbf{Residual vector quantization.} Instead of quantizing $z$ with a single codebook, RVQ progressively represents the remaining residual. Starting from $r^{(0)}=z$, the $k$-th codebook performs
\begin{equation}
q^{(k)}
=
\arg\min_{e\in\mathcal{C}^{(k)}}
\left\|r^{(k-1)}-e\right\|_2^2,
\qquad
r^{(k)}
=
r^{(k-1)}-\operatorname{sg}(q^{(k)}),
\label{eq:rvq_stage}
\end{equation}
and the final quantized representation is $Q(z)=\sum_{k=1}^{K}q^{(k)}$. Each stage refines the approximation left by the preceding one, so the representation capacity grows with $K$ instead of being fixed by a single codebook. Quantizers of this form were introduced for neural audio coding~\cite{zeghidour2022soundstream} and later used for motion synthesis~\cite{guo2024momask}. We exploit the same property to obtain a distillation space compact enough to regularize GM yet expressive enough for the motions it represents.

\textbf{Pretraining objective.} Since the nearest-codeword selection is not differentiable, a straight-through estimator passes gradients from the decoder to the continuous latent representation during training. The RVQ-VAE is optimized using $\mathcal{L}_{\mathrm{RVQ}}=\mathcal{L}_{\mathrm{rec}}+\lambda_v\mathcal{L}_{\mathrm{vel}}^{\mathrm{rec}}+\mathcal{L}_{\mathrm{codebook}}+\mathcal{L}_{\mathrm{commit}}+\lambda_{\mathrm{KL}}(t)\mathcal{L}_{\mathrm{KL}}$, where $\mathcal{L}_{\mathrm{rec}}$ reconstructs the motion features, $\mathcal{L}_{\mathrm{vel}}^{\mathrm{rec}}$ reconstructs their first-order differences with weight $\lambda_v$ (not to be confused with the distillation term $\mathcal{L}_{\mathrm{vel}}$ in Eq.~(\ref{eq:final_loss})), $\mathcal{L}_{\mathrm{codebook}}$ and $\mathcal{L}_{\mathrm{commit}}$ are the codebook and commitment losses of VQ-VAE~\cite{oord2017vqvae} summed over the $K$ stages, and $\lambda_{\mathrm{KL}}(t)$ is an annealed KL weight. The KL term pulls the pre-quantization latent toward a standard Gaussian, bounding the scale of the space in which distillation later operates. After pretraining, the encoder, codebooks, and decoder are frozen.

\begin{table*}[!t]
\centering
\caption{Comparisons at the $1\times$ synthetic-data budget on Human3.6M, CMU, and 3DPW. Results are MPJPE (mm) at frames \#2/\#4/\#8/\#10/\#14, given as mean$\pm$std over five runs, and Mean5 is the average of the five horizon-wise means. Best compressed-set results are in bold, and Full is trained under the same fixed budget and serves as a reference.}
\label{tab:main_1x}
{
\UnifiedTableStyle
\begin{tabular*}{0.96\textwidth}{
@{\extracolsep{\fill}}
llccccc>{\columncolor{oursbg}}c
@{}
}
\toprule
Dataset & Method & \#2 & \#4 & \#8 & \#10 & \#14 & Mean5 \\
\midrule
\multicolumn{8}{c}{(a) Simple} \\
\midrule
\multirow{4}{*}{H36M}
& Full & $7.48{\pm}0.37$ & $15.14{\pm}0.67$ & $31.72{\pm}1.08$ & $40.72{\pm}1.19$ & $58.98{\pm}1.86$ & 30.81 \\
\cmidrule(lr){2-8}
& Random & $17.72{\pm}1.22$ & $34.88{\pm}3.17$ & $68.34{\pm}4.71$ & $82.98{\pm}6.00$ & $107.94{\pm}7.66$ & 62.37 \\
& GM & $45.84{\pm}10.42$ & $56.12{\pm}16.20$ & $92.16{\pm}18.85$ & $114.08{\pm}12.39$ & $119.52{\pm}14.32$ & 85.54 \\
& Ours & $\boldsymbol{13.78{\pm}0.60}$ & $\boldsymbol{26.50{\pm}0.89}$ & $\boldsymbol{47.38{\pm}1.27}$ & $\boldsymbol{57.04{\pm}1.81}$ & $\boldsymbol{73.94{\pm}1.13}$ & \textbf{43.73} \\
\midrule
\multirow{4}{*}{CMU}
& Full & $11.82{\pm}1.27$ & $23.09{\pm}2.26$ & $47.62{\pm}3.08$ & $60.56{\pm}2.80$ & $87.90{\pm}2.93$ & 46.20 \\
\cmidrule(lr){2-8}
& Random & $30.54{\pm}2.63$ & $58.65{\pm}6.83$ & $111.07{\pm}12.96$ & $131.82{\pm}14.65$ & $176.85{\pm}13.34$ & 101.79 \\
& GM & $32.36{\pm}5.48$ & $51.58{\pm}7.41$ & $88.38{\pm}11.10$ & $108.54{\pm}15.44$ & $133.25{\pm}12.93$ & 82.82 \\
& Ours & $\boldsymbol{21.41{\pm}0.99}$ & $\boldsymbol{41.47{\pm}1.31}$ & $\boldsymbol{77.03{\pm}2.49}$ & $\boldsymbol{92.56{\pm}3.55}$ & $\boldsymbol{118.73{\pm}5.00}$ & \textbf{70.24} \\
\midrule
\multirow{4}{*}{3DPW}
& Full & $18.83{\pm}2.77$ & $34.94{\pm}3.28$ & $67.34{\pm}7.44$ & $79.32{\pm}8.14$ & $100.85{\pm}13.81$ & 60.26 \\
\cmidrule(lr){2-8}
& Random & $38.18{\pm}11.20$ & $68.57{\pm}19.25$ & $112.72{\pm}30.57$ & $130.32{\pm}31.00$ & $155.73{\pm}35.67$ & 101.10 \\
& GM & $31.83{\pm}5.27$ & $49.60{\pm}5.13$ & $82.22{\pm}11.43$ & $96.23{\pm}10.24$ & $116.63{\pm}12.05$ & 75.30 \\
& Ours & $\boldsymbol{22.82{\pm}0.96}$ & $\boldsymbol{42.68{\pm}2.65}$ & $\boldsymbol{74.68{\pm}3.29}$ & $\boldsymbol{86.56{\pm}3.71}$ & $\boldsymbol{104.85{\pm}4.14}$ & \textbf{66.32} \\
\midrule
\multicolumn{8}{c}{(b) DLinear} \\
\midrule
\multirow{4}{*}{H36M}
& Full & $14.46{\pm}2.53$ & $30.16{\pm}3.45$ & $64.54{\pm}8.30$ & $81.52{\pm}10.22$ & $112.02{\pm}14.38$ & 60.54 \\
\cmidrule(lr){2-8}
& Random & $17.92{\pm}7.18$ & $34.70{\pm}8.10$ & $66.84{\pm}8.25$ & $78.42{\pm}10.22$ & $107.58{\pm}13.53$ & 61.09 \\
& GM & $19.88{\pm}4.81$ & $35.24{\pm}4.88$ & $60.28{\pm}3.45$ & $70.20{\pm}2.61$ & $85.04{\pm}3.15$ & 54.13 \\
& Ours & $\boldsymbol{11.44{\pm}1.05}$ & $\boldsymbol{24.28{\pm}2.10}$ & $\boldsymbol{47.62{\pm}4.95}$ & $\boldsymbol{57.96{\pm}5.57}$ & $\boldsymbol{80.16{\pm}9.81}$ & \textbf{44.29} \\
\midrule
\multirow{4}{*}{CMU}
& Full & $19.33{\pm}0.36$ & $36.89{\pm}0.61$ & $71.49{\pm}0.75$ & $86.64{\pm}0.66$ & $112.29{\pm}1.12$ & 65.33 \\
\cmidrule(lr){2-8}
& Random & $16.93{\pm}1.67$ & $35.93{\pm}2.86$ & $75.14{\pm}4.74$ & $92.80{\pm}5.25$ & $123.27{\pm}6.81$ & 68.81 \\
& GM & $20.76{\pm}2.67$ & $39.79{\pm}3.52$ & $74.44{\pm}2.74$ & $89.37{\pm}2.06$ & $\boldsymbol{111.79{\pm}1.09}$ & 67.23 \\
& Ours & $\boldsymbol{12.85{\pm}1.35}$ & $\boldsymbol{28.96{\pm}3.07}$ & $\boldsymbol{67.33{\pm}7.36}$ & $\boldsymbol{86.11{\pm}9.60}$ & $118.18{\pm}13.53$ & \textbf{62.69} \\
\midrule
\multirow{4}{*}{3DPW}
& Full & $19.05{\pm}0.29$ & $37.51{\pm}0.62$ & $69.79{\pm}0.58$ & $81.86{\pm}0.74$ & $99.63{\pm}1.05$ & 61.57 \\
\cmidrule(lr){2-8}
& Random & $19.69{\pm}3.56$ & $38.87{\pm}7.79$ & $74.84{\pm}12.01$ & $88.53{\pm}13.36$ & $109.45{\pm}18.92$ & 66.28 \\
& GM & $18.83{\pm}1.74$ & $37.83{\pm}3.69$ & $70.85{\pm}4.21$ & $\boldsymbol{83.63{\pm}4.26}$ & $\boldsymbol{102.82{\pm}5.60}$ & \textbf{62.79} \\
& Ours & $\boldsymbol{17.96{\pm}0.81}$ & $\boldsymbol{36.05{\pm}1.52}$ & $\boldsymbol{70.65{\pm}3.38}$ & $84.70{\pm}3.91$ & $107.12{\pm}5.31$ & 63.30 \\
\bottomrule
\end{tabular*}
}
\end{table*}

\begin{figure*}[t]
    \centering
    \includegraphics[width=1\textwidth]{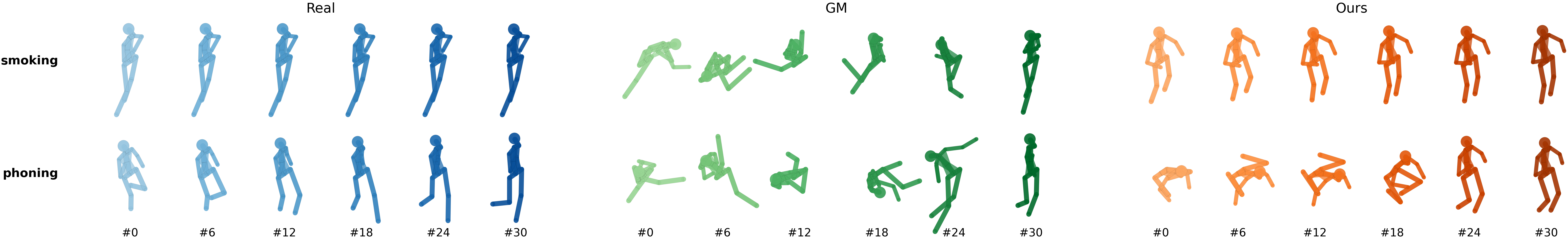}
    \caption{Synthetic training trajectories (not predictions) initialized from two Human3.6M sequences; the left column shows the initialization, and frame indices \#0--\#30 index the 50-frame trajectory rather than prediction horizons. The distilled motions need not retain the initializing action. In the examples shown, direct GM produces implausible joint configurations and abrupt transitions, whereas motions decoded from the latent bank stay within the decoder output space.}
    \label{fig:qualitative}
\end{figure*}

\begin{figure*}[t]
    \centering
    \includegraphics[width=1\textwidth]{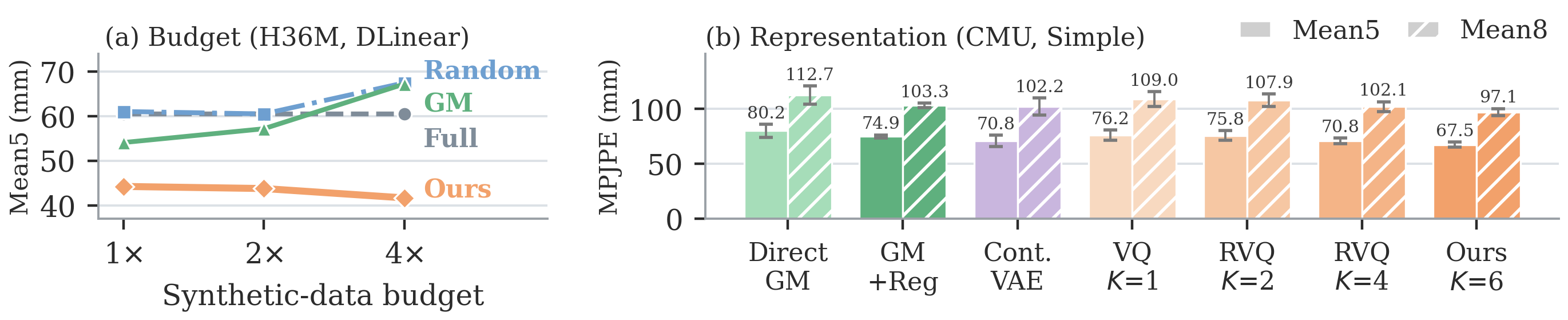}
    \caption{Ablation studies. (a) Mean5 MPJPE on Human3.6M with DLinear as the budget grows, where Full is budget-independent and trained for the same 4000 updates. (b) Mean5 and Mean8 MPJPE on CMU with Simple at the $2\times$ budget. GM+Reg is direct GM trained with the objective of Eq.~(\ref{eq:final_loss}), $K=1$ is a single-stage VQ variant, and error bars are standard deviations over five runs.}
    \label{fig:ablation}
\end{figure*}

\subsection{Latent Dataset Distillation}

The framework that performs distillation over the pretrained latent space is illustrated in Fig.~\ref{fig:method}.

\textbf{Latent-bank initialization.} For a budget of $M$ synthetic trajectories, we first select real 50-frame segments $\{X_i^{(0)}\}_{i=1}^{M}$ and initialize a latent bank $\mathcal{H}=\{h_i\}_{i=1}^{M}$ using the encoder mean branch, $h_i^{(0)}=E_{\mu}(X_i^{(0)})$. A trajectory $\widetilde{X}_i=D_{\psi}(Q(h_i))$ is then decoded from each slot with the frozen RVQ and decoder, so distillation searches over $\mathcal{H}$ rather than over all motion values.

\textbf{Gradient flow through the frozen quantizer.} Because nearest-codeword selection is piecewise constant in $h_i$, its gradient is zero almost everywhere and gives no optimization signal. During distillation, we therefore apply the same straight-through estimator to the frozen quantizer,
\begin{equation}
Q_{\mathrm{ST}}(h)=h+\operatorname{sg}\!\left(Q(h)-h\right),
\label{eq:ste}
\end{equation}
so that the forward pass decodes $Q(h)$ while the gradient of the distillation objective reaches $h$ directly.

\textbf{Motion-space regularization.} At each iteration, a batch of 50-frame real segments is sampled from the full training pool, and synthetic segments are decoded from randomly selected latent slots, so the sequence or action used for initialization does not determine the real--synthetic pairing. For each pair, 35-frame windows $W_r=[P_r,F_r]$ and $W_s=[P_s,F_s]$ are extracted with the same random temporal offset and passed to the shared backbone to compute $\mathcal{L}_{\mathrm{GM}}$. We further regularize the decoded motion using $\mathcal{L}_{\mathrm{pose}}=\operatorname{MSE}(W_s,W_r)$ and $\mathcal{L}_{\mathrm{vel}}=\operatorname{MSE}(\Delta W_s,\Delta W_r)$, where $\Delta W_t=W_{t+1}-W_t$. These regularizers do not assume semantic correspondence between paired windows. Under random pairing, their expectation splits into a term minimized at the mean real window and a term independent of $W_s$, so they act as a mean-shrinkage regularizer that anchors the decoded motion to the bulk of the real distribution. Note that $\mathcal{L}_{\mathrm{vel}}$ operates on differences of rotation parameters rather than physical angular velocities.

\textbf{Distillation objective.} The complete objective is
\begin{equation}
\mathcal{L}_{\mathrm{distill}}
=
\lambda_{\mathrm{GM}}\mathcal{L}_{\mathrm{GM}}
+
\lambda_{\mathrm{vel}}\mathcal{L}_{\mathrm{vel}}
+
\lambda_{\mathrm{pose}}\mathcal{L}_{\mathrm{pose}}.
\label{eq:final_loss}
\end{equation}
Only $\mathcal{H}$ is updated, while the RVQ-VAE remains fixed and the backbone provides gradient-matching signals. After optimization, all latent slots are decoded once into 50-frame sequences, yielding a standard synthetic dataset that trains predictors without the RVQ-VAE.

\section{Experiments}
\label{sec:experiments}

\subsection{Experimental Setup}

\textbf{Datasets and Evaluation Metrics.} We evaluate the framework on Human3.6M~\cite{ionescu2014h36m}, the CMU Graphics Lab Motion Capture Database~\cite{cmumocap}, and 3DPW~\cite{marcard2018_3dpw}. Human3.6M is represented by 99-D exponential maps, CMU by 70-D normalized features, and 3DPW by 72-D SMPL axis-angle features. Prediction quality is measured by mean per-joint position error (MPJPE, mm) on 3D joint positions recovered from each representation, where a lower value is better. Models observe 25 frames and predict the next 10; at test time the predicted block is appended to the observation window, and the predictor is reapplied until 25 future frames are produced, so horizons beyond \#10 come from rollout. We report frames \#2/\#4/\#8/\#10/\#14, corresponding to 80/160/320/400/560~ms at 25~fps. Mean5 averages these horizons, and Mean8 adds \#18/\#22/\#25 (720/880/1000~ms).

\textbf{Baselines and Budgets.} We compare Full-data training, Random subset selection, direct GM, and the proposed method. Direct GM optimizes the same initialized segments in the feature space using only the GM objective, with the same budget, backbone, iterations, and learning rate as Ours; the ablation also reports direct GM trained with the objective of Eq.~(\ref{eq:final_loss}), denoted GM+Reg. The backbone used as the gradient probe is randomly initialized and kept fixed within each run. Adapting temporal DD methods~\cite{miao2024timedc,hong2026hdt,kwon2026stemdist} to articulated motion is left for future work. The budget is the number $M$ of synthetic trajectories, each of which yields 16 overlapping, hence non-independent, 35-frame training windows. The $1\times$/$2\times$/$4\times$ budgets correspond to 5\%/10\%/20\% of the real training sequences on Human3.6M, 10\%/20\%/40\% on 3DPW, and 1/2/4 trajectories per action class on CMU, with identical $M$ for Random, GM, and Ours.

\textbf{Implementation Details.} We use a lightweight backbone based on SiMLPe~\cite{guo2023simlpe}, denoted Simple, and DLinear~\cite{zeng2023dlinear}. The RVQ-VAE uses a 64-D latent space, six residual stages, and 128 entries per codebook, pretrained for 4000 iterations with a learning rate of $2\times10^{-4}$. We set $\lambda_v=1$ and warm up $\lambda_{\mathrm{KL}}(t)=10^{-4}\min(1,t/1000)$ over the first 1000 iterations. The latent bank is distilled for 2500 iterations with Adam, learning rate $10^{-2}$, and batch size 16, with $\lambda_{\mathrm{GM}}=10$ and $\lambda_{\mathrm{pose}}=\lambda_{\mathrm{vel}}=1.5$. All downstream models, including Full, use the same setting of 4000 iterations, learning rate $3\times10^{-4}$, and weight decay $10^{-4}$, so Full is a reference under this protocol rather than a converged upper bound. We report results averaged over five random seeds.

\subsection{Evaluation Results}

\textbf{Comparisons at the Fixed Budget.} Table~\ref{tab:main_1x} compares the proposed method with Full, Random, and direct GM at the $1\times$ budget. Ours achieves the lowest Mean5 MPJPE in five of the six dataset--backbone combinations, cutting the error of direct GM by 48.9\% on Human3.6M with Simple and by 6.8\% to 18.2\% on four pairs, while 3DPW with DLinear is 0.8\% worse. Over the 30 horizon-level comparisons, Ours leads direct GM in 27 and Random in all 30. 
Direct GM is unstable on Human3.6M with Simple, where its error exceeds that of Random with a large variance. 
Its distilled motions contain implausible joint configurations (Fig.~\ref{fig:qualitative}).
Ours falls below Full on several DLinear settings, where Full is trained for the same 4000 updates and has not converged.

\textbf{Ablation on the Data Budget.} Fig.~\ref{fig:ablation}(a) varies the synthetic-data budget on Human3.6M with DLinear. Ours achieves 44.29, 43.82, and 41.72 Mean5 at the three budgets, beating direct GM and Random at every budget. Random and direct GM instead degrade at $4\times$, where the fixed downstream budget gives each trajectory fewer passes over training.

\begin{table}[t]
\centering
\caption{Cross-backbone transfer at the $2\times$ budget. Each cell is the Mean5 MPJPE (mm) of the target backbone trained on the set distilled with the source backbone (S: Simple, D: DLinear). Parentheses give the target backbone trained on a set distilled with the target backbone itself.}
\label{tab:cross_backbone}
\UnifiedTableStyle
\setlength{\tabcolsep}{2pt}
\begin{tabular*}{\columnwidth}{@{\extracolsep{\fill}}lc>{\columncolor{oursbg}}c@{}}
\toprule
Transfer & GM & Ours \\
\midrule
CMU S$\rightarrow$D & $70.40{\pm}3.48$ (68.37) & $\mathbf{58.45{\pm}2.61}$ (63.32) \\
CMU D$\rightarrow$S & $100.34{\pm}7.41$ (80.21) & $\mathbf{90.98{\pm}4.35}$ (67.51) \\
H36M S$\rightarrow$D & $66.12{\pm}14.37$ (57.22) & $\mathbf{47.30{\pm}4.05}$ (43.82) \\
H36M D$\rightarrow$S & $102.16{\pm}14.05$ (70.97) & $\mathbf{78.41{\pm}9.18}$ (43.89) \\
\bottomrule
\end{tabular*}
\end{table}

\textbf{Cross-Backbone Transfer.} Table~\ref{tab:cross_backbone} asks whether a set distilled with one backbone stays useful for another, and Ours yields lower transfer error than direct GM in all four directions. The gain is clearest for Simple$\rightarrow$DLinear, where the transferred data stays close to the in-backbone result and on CMU outperforms it. The distilled sets thus retain cross-backbone utility in absolute accuracy, while the relative degradation from in-backbone to transferred training is larger for Ours than for direct GM in two of the four directions.

\textbf{Ablation on the Latent Representation.} Fig.~\ref{fig:ablation}(b) compares latent representations at the $2\times$ budget. Adding the regularizers of Eq.~(\ref{eq:final_loss}) to direct GM lowers Mean5 from 80.2 to 74.9, and all latent variants use this same objective, so the remaining differences reflect the representation alone. Relative to GM+Reg, the continuous VAE latent space lowers Mean5 to 70.8, a single-stage VQ variant is worse than the continuous space at 76.2, and increasing the residual depth recovers the loss, with $K=6$ reaching 67.5 Mean5 and 97.1 Mean8, 9.9\% and 6.0\% below GM+Reg and with the smallest spread among all variants. A larger $K$ also enlarges the codebooks, so the gain in residual depth includes the added capacity.

\textbf{Qualitative Comparison.} Fig.~\ref{fig:qualitative} shows synthetic training trajectories, not predictions. Since pairing ignores the initializing sequence, the distilled motions need not retain the initializing action. In the examples shown, direct GM produces implausible joint configurations and abrupt transitions, whereas motions decoded from the latent bank stay within the decoder output space.

\section{Conclusion}
We have presented a latent dataset distillation framework for human motion prediction. Rather than optimizing motion features directly, it distills in the pre-quantization latent space of a pretrained RVQ-VAE and updates only a latent bank through the frozen quantizer and decoder, so gradient matching is regularized by a learned motion prior. On Human3.6M, CMU, and 3DPW with two backbones, the distilled sets outperform direct gradient matching and random subsets under a fixed downstream budget, and residual quantization improves over single-stage quantization among latent variants sharing the same objective. The distilled sets also remain useful when transferred to a backbone other than the one used for distillation.

\clearpage
\bibliographystyle{IEEEbib}
\bibliography{reference}
\end{document}